\documentclass[%
 aip,
cp,  
 amsmath,amssymb,
 reprint,%
]{revtex4-2}
\usepackage{subcaption}
\usepackage{graphicx}
\usepackage{dcolumn}
\usepackage{bm}

\usepackage[utf8]{inputenc}
\usepackage[T1]{fontenc}
\usepackage{mathptmx} 
\usepackage{subcaption}
\usepackage{multirow}
\begin{document}

\title{Improved Defect Detection and Classification Method for Advanced IC Nodes by Using Slicing Aided Hyper Inference with Refinement Strategy}

\affiliation{Ghent University, Ghent, 9000, Belgium}
\affiliation{imec, Leuven, 3001, Belgium}
\affiliation{SCREEN SPE Germany GmbH, Ismaning, 85737, Germany}
\author{Vic De Ridder}
\altaffiliation{These authors contributed equally}
\email{vic.deridder.ext@imec.be}
\affiliation{Ghent University, Ghent, 9000, Belgium}
\affiliation{imec, Leuven, 3001, Belgium}
\author{Bappaditya Dey}
\altaffiliation{These authors contributed equally}
\email{bappaditya.dey@imec.be}
\affiliation{imec, Leuven, 3001, Belgium}
\author{Victor Blanco}
\affiliation{imec, Leuven, 3001, Belgium}
\author{Sandip Halder}
\altaffiliation{This research was conducted during Sandip Halder's  tenure at imec}
\affiliation{SCREEN SPE Germany GmbH, Ismaning, 85737, Germany}
\author{Bartel Van Waeyenberge}
\affiliation{Ghent University, Ghent, 9000, Belgium}


\begin{abstract}
In semiconductor manufacturing, lithography has often been the manufacturing step defining the smallest possible pattern dimensions. In recent years, progress has been made towards high-NA (Numerical Aperture) EUVL (Extreme-Ultraviolet-Lithography) paradigm, which promises to advance pattern shrinking (2 nm node and beyond). However, a significant increase in stochastic defects and the complexity of defect detection becomes more pronounced with high-NA. Present defect inspection techniques (both non-machine learning and machine learning based), fail to achieve satisfactory performance at high-NA dimensions.
In this work, we investigate the use of the Slicing Aided Hyper Inference (SAHI) framework for improving upon current techniques. Using SAHI, inference is performed on size-increased slices of the SEM images. This leads to the object detector's receptive field being more effective in capturing small defect instances. First, the performance on previously investigated semiconductor datasets is benchmarked across various configurations, and the SAHI approach is demonstrated to substantially enhance the detection of small defects, by $\sim$ 2x. Afterwards, we also demonstrated application of SAHI leads to flawless detection rates on a new test dataset, with scenarios not encountered during training, whereas previous trained models failed. Finally, we formulate an extension of SAHI that does not significantly reduce true-positive predictions while eliminating false-positive predictions.
\end{abstract}
\maketitle

\section{Introduction}

As Machine Learning (ML) gets increasingly adopted in different domains such as drug discovery, self-driving cars, financial risk analysis and many other extensive computational applications, importance of developing faster and energy-efficient integrated circuits (IC's) continues to grow. In producing an IC, lithography is a crucial step, involving deposition of a photoresist on the raw silicon wafer and then exposing certain parts of this photoresist to lightwaves such that an intended geometrical pattern gets printed. Therefore, the smallest possible, printable pattern dimension is largely limited by lithography, which itself is limited in large part by light-wavelength $\lambda$ and numerical aperture (NA) of the lens used, following Rayleigh criterion \cite{rayleigh}. In the last decade, production of smaller device features has been enabled by smaller wavelengths through Extreme Ultra Violet Lithography (EUVL), which has $\lambda=13.5$nm, NA=0.33. A next step being developed towards enabling further pattern shrinkage (beyond $\sim$ 2nm node technology) is high-NA EUVL (NA=0.55).
With the deployment of high-NA EUVL, improved metrology and nano-scale defect inspection is crucial at various R\&D stages of semiconductor manufacturing. 

Since rule-based, classical defect detection approaches failed at these advanced nodes \cite{rulesfail}, Machine Learning (ML-) based methods have been developed as an alternative \cite{reviewenrique}. Nevertheless, the reliability and adaptability of these machine learning models have not been fully verified for high-volume manufacturing (HVM). Several factors to be considered, are i) limited availability of some stochastic defect patterns/types in annotated training dataset, and ii) ambiguity in defining precise pixel extent (foreground/background) for some defect classes/instances. Additionally, due to the reduced depth of focus, thickness of photoresists for high-NA EUVL is lower compared to previous lithography approaches (EUVL), leading to increased noise and lower contrast in SEM (Scanning Electron Microscope) imaging, and thus even more challenging defect inspection conditions are emerging towards facilitating high-NA enabled lithography.

In this research work, we investigate using the Slicing Aided Hyper Inference (SAHI) framework \cite{sahi} to improve SEM-based defect detection of resist wafers at EUVL process pitch-scales and beyond. We evaluate the performance of SAHI on two semiconductor SEM-based defect inspection datasets with different patterns: Hexagonal Contact Hole Directed-Self-Assembly (HEXCH DSA) and line-space (LS). SAHI is a model-agnostic inference framework, and can thus be appended to various ML-based ADCD (Automated Defect Classification and Detection) frameworks. Our main contributions are: 

(i) We investigated SAHI-based inference framework for two different YOLO models/architecture variants and different configurations on two semiconductor SEM datasets: Line-space and Hexagonal Contact-Hole Arrays, to demonstrate defect detection performance improvement, specifically for challenging nano-scale defect types.

(ii) We demonstrated SAHI-based framework resulted in flawless detection rates on a novel test dataset, encompassing scenarios that were not encountered during training, in contrast to the previous trained models, which experienced significant failures.

(iii) Finally, we propose a new strategy appended to SAHI-framework, to refine model predictions towards eliminating false-positive predictions.

\section{Previous Work}
\subsection{Semiconductor Defect Inspection}
Ref.\cite{rulesfail}, attempted defect inspection with traditional, rule-based algorithms on semiconductor wafers in line with high-NA EUVL requirements, and concluded that as resist thickness shrinks, rule-based methods fail to detect any defects. Other, non ML-methods have already been proposed for EUVL, such as \cite{statapproach}, which is a statistical approach that uses reference images without defects. However, the reliance on reference images has two disadvantages. First, a proper reference image needs to be available, which is not necessarily the case, especially when performing defect inspection on more complex, irregular patterns. Secondly, it relies on precise, near perfect alignment between the reference image and inspected images, which is extremely difficult for nano-scale patterns due to process window variation \cite{prwv}. Moreover, the defect detection performance of this type of approach on aggressive pitches ($\sim$2 nm and beyond) is significantly affected by higher noise levels \cite{ourretina}.

To tackle these challenges, it has been proposed to use advanced ML-based object detection frameworks for defect classification and detection \cite{ourretina}. This approach does not rely on any reference images and can learn to detect challenging stochastic defects even in the presence of significant noise levels and low contrast, provided it has sufficient annotated data. Ref.\cite{augm} proposed using YOLOv3 \cite{YOLOv3} and compared different training data strategies, concluding that training on mixture of real and simulated SEM images leads to the best performance. These simulated SEM images increase total training data, and do not require manual annotation. However, as of now there is no method of accurately simulating SEM images with line-width roughness or noise parameters identical to those obtained from real SEM imaging tools. With high-NA EUVL, stochastic noise (and low Signal-to-Noise-Ratio) will be one of the major concerns that need to be tackled,  thus using simulated data in training does not seem to be a viable strategy for advanced defect inspection in HVM (High-Volume-Manufacturing) using thin-resists. 

While various object detection architectures have been explored for defect inspection task \cite{reviewenrique, ourYOLOv5, ourYOLOv8}, this is a time-consuming procedure and its results may not necessarily generalize to different semiconductor datasets, which is why different methods for improving semiconductor defect inspection performance need to be further investigated. Hence, this research work aims to investigate the object detection inference framework SAHI \cite{sahi}, which was originally proposed for improving small object detection and can be added to any defect inspection pipeline that uses ML-based object detectors.
\begin{figure}[t]
    \centering
    \includegraphics[width=\textwidth]{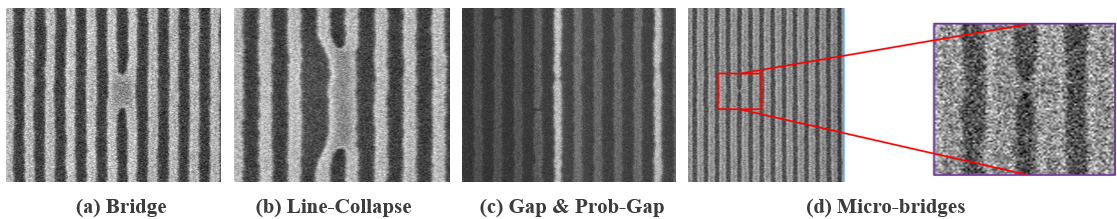}
    \caption{Example defect types in the line-space dataset}
    \label{adi_ex}
\end{figure}

\begin{figure}[t]
  \centering
  \begin{subfigure}{0.2\textwidth}
    \includegraphics[width=\linewidth]{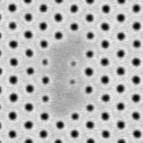}
    \caption{Closed Patch}
  \end{subfigure}
  \begin{subfigure}{0.2\textwidth}
    \includegraphics[width=\linewidth]{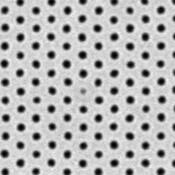}
    \caption{Partially Closed Hole}
  \end{subfigure}
    \begin{subfigure}{0.2\textwidth}
    \includegraphics[width=\linewidth]{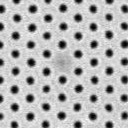}
    \caption{Missing Hole}
  \end{subfigure}
  \caption{Examples defect types in HEXCH DSA dataset}
  \label{dsa_ex}
\end{figure}

\subsection{Small Object Detection with SAHI}

In most object detection frameworks, predictions are obtained by passing the image as a whole to the model. 
By contrast, the SAHI inference framework \cite{sahi} slices the image into overlapping patches and increases the size of these slices before passing them to the model, which in turn helps to significantly improve performance of detection models on smaller objects.
Public object detection datasets such as COCO contain both small and large instances of each object type. For these object classes, the smaller instances (also in a size increased slice using SAHI) share feature similarity with larger instances, which object detectors already trained with. Therefore, no additional fine-tuning was required to improve the detection performance. 

One popular method for increasing detection accuracy of small objects is Feature Pyramid Networks (FPN) \cite{retinanet}, which integrates feature maps of various resolutions into a unified object detection head. RetinaNet based ADCD framework (which uses FPN) was proposed and investigated on LS (ADI) dataset \cite{ourretina} and also compared against several other benchmark models \cite{reviewenrique}. However, though performance was significantly improved for several (comparatively larger)defect classes, for smaller defect classes/instances it was negligible.

While other ML architectures have been proposed for small object detection \cite{reviewsmallobject}, we decide to investigate SAHI on private semiconductor datasets (mainly for above discussed challenging smaller defect classes and instances) for three main reasons: i) SAHI is model-agnostic, which enables integration into already existing ADCD pipelines. 
ii) SAHI-based inference strategy increases the computational time, however, for semiconductor industry, the primary concern is acquiring best dataset to feed to ML models (as relevant data is not just rare and noisy, but also extremely expensive to get).
iii) towards enabling data-centric ML-based improved defect inspection, which means rather than investigating different models (to increase performance time-to-time), to investigate and improve dataset itself (generally by correcting mislabeled/unannotated classes and instances).
\begin{figure}[h]
    \centering
    \includegraphics[width=\linewidth]{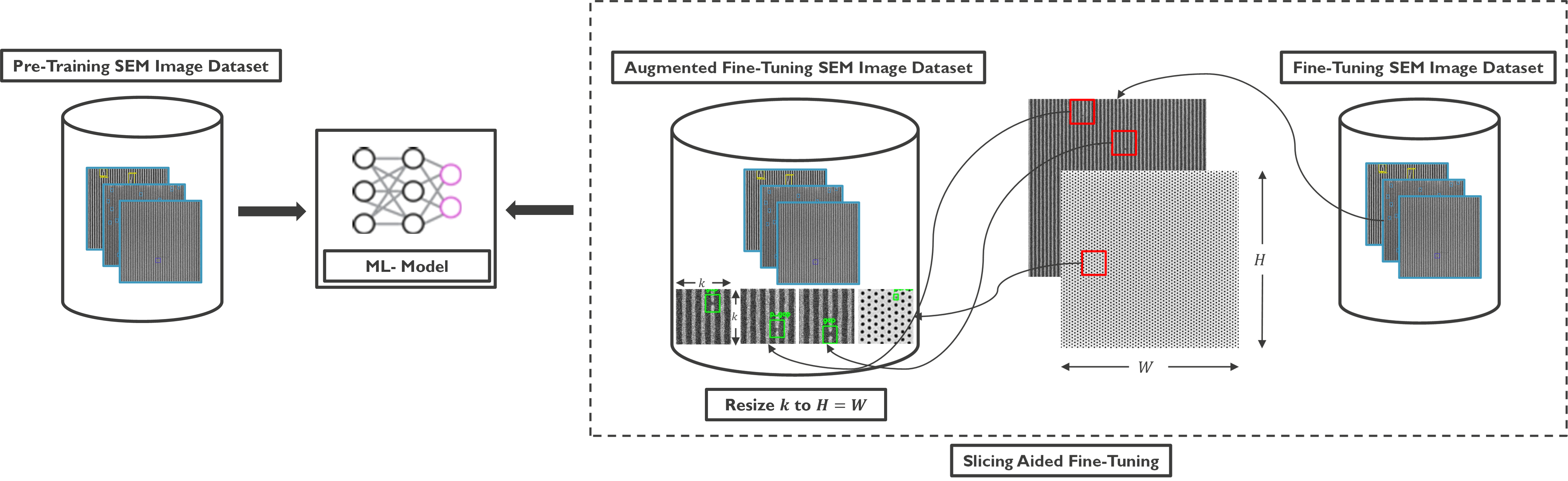}
    \caption{Illustration of proposed semiconductor ADCD framework based on SAHI \cite{sahi}}
    \label{sahi_illustration}
\end{figure}

\section{Datasets}\label{dataset}

In this work, performance was investigated on two different semiconductor SEM datasets: LS (ADI) \cite{ourretina} and HEXCH DSA \cite{ourYOLOv8}. In this study, these two datasets are reused from \cite{ourretina, ourYOLOv8} to demonstrate improvement over previous benchmarking performance. No synthetic images were used. All encountered defects were stochastic in nature and hand-labeled by expert annotators.
The first dataset consists of line-space patterned resist wafer SEM images (ADI) with possible defect types as shown in fig.\ref{adi_ex}: probable gap (pgap), gap, bridge, microbridge, and line collapse. In previous works \cite{ourYOLOv5, ourYOLOv7}, mAP was significantly lower on the pgap defect type compared to all others. The probable cause can be hypothesized as the interplay of following two problems: i) pgap has the smallest defect pixel area, thus it is hard for the model to learn to differentiate between the corresponding background pixels and gap defect instance pixels, and detect the precise pgap features. ii) Relatively few pgap instances are present inside the training dataset (table.\ref{adidataset}). Hence this research work will put an emphasis of SAHI's impact on pgap detection. In the HEXCH DSA dataset, printed patterns are hexagonal Contact Hole (HEXCH) arrays, with three different defect types shown in fig.\ref{dsa_ex}: closed patch (cp), missing hole (mh), partially closed hole (pch).

\begin{table}[t]
\renewcommand{\arraystretch}{1.3}
\caption{\textsc{Line-space (ADI) Dataset Distribution and Statistics}}
\label{adidataset}
\centering
\begin{tabular}{|c||c||c||c||c|}
\hline
  &&\textbf{Training}  & \textbf{Validation}&\textbf{Test} \\
\hline
\textbf{Total Images} &  &1053 & 117 & 154\\
\hline
\textbf{Class name} & \textbf{Abbreviation}&\textbf{Training} & \textbf{Validation} & \textbf{Test} \\
\hline
Gap & gap &1046 & 156 &174\\
\hline
Probable Gap &pgap  & 315 & 49 & 54\\
\hline
Bridge&  bridge&238 & 19 &17\\
\hline
Microbridge&mb &380 & 47 &78\\
\hline
Line Collapse &lc &555 & 66 & 76\\
\hline
\textbf{Total Instances} &  &\textbf{2529} & \textbf{337} & \textbf{399}\\
\hline
\end{tabular}
\end{table}

\begin{table}[t]
\renewcommand{\arraystretch}{1.3}
\caption{\textsc{Hexagonal DSA Dataset Distribution and Statistics}}
\label{tdataset}
\centering
\begin{tabular}{|c||c||c||c||c|}
\hline
  & &\textbf{Training} & \textbf{Validation} & \textbf{Test}\\
\hline
\textbf{Total Images} & &174  &34  & 42\\
\hline
\textbf{Class name} & \textbf{Abbreviation}&\textbf{Training} & \textbf{Validation} & \textbf{Test} \\
\hline
Missing Hole&mh & 30  &  8&5\\
\hline
Partially Closed Hole&pch  &94  & 16& 23\\
\hline
Closed Patch&cp & 74  & 13 &19\\
\hline
\textbf{Total Instances}& &  \textbf{198} & \textbf{37} & \textbf{47}\\
\hline
\end{tabular}
\end{table}

\section{Proposed Methodology}

\subsection{Training}\label{init}
To evaluate the application of SAHI on ADCD performance, following architectures are trained on both HEXCH DSA and LS (ADI) dataset from COCO \cite{coco} pretrained weights: YOLOv8 (n,s,m,l,x) and YOLOv5 (n,m,x).
On each dataset, models are trained for a maximum of 200 epochs, with early stopping enabled and batch size of 32. For each model, application of SAHI is investigated for the variation(s) that achieve(s) the highest mAP by normal inference (without SAHI), as demonstrated in tables \ref{adiresultss} and \ref{dsaresultss} respectively.


While the SAHI framework could improve performance on the datasets considered in the original work \cite{sahi} without requiring fine-tuning, the same applied on the investigated semiconductor datasets showed problematic results. The considered semiconductor defects have limited intra-class variations in pixel area. Thus, when model is only trained on full images, it never encountered defects at the size they would be in the size-increased slices, and was not able to make correct predictions within SAHI framework.

Hence, models are finetuned on sliced datasets before performing SAHI-based inference. The finetuning strategy is adopted from the original work \cite{sahi}, where it was first proposed for increased performance. Sliced datasets are made for slice-sizes of 128, 256, and 512 at an overlap ratio of 0.5 and with the restraint that at least one defect is contained entirely within that slice. Models are finetuned for 50 epochs on each of these sliced datasets, with the same training hyperparameters as those of the normal training. Our proposed ADCD framework is illustrated in fig.\ref{sahi_illustration}.

\subsection{Comparison and Metrics}
In initial experimentation, average-precision of SAHI-based predictions was significantly lower than baseline, while average-recall was better than baseline. Manual visual inspection of the predictions on validation data showed that SAHI-based predictions included a lot of correct predictions, which were not included in the annotations. Thus, manual inspection is performed on both without SAHI and SAHI-based inference results when comparing the two methods against each other. The number of true- and false-positives (both according to manual evaluation of each prediction) is used for comparisons.  During the comparisons, a confidence threshold of 0.25 is used, as this was found to be optimal for both standard and SAHI-based inference. All SAHI-based methods are evaluated for an overlap ratio of 0.1 between patches.

\begin{figure}
    \centering
    \includegraphics[scale=0.6]{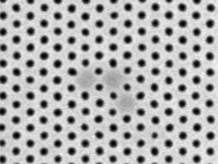}
    \caption{Example of DSA defect scenario where conventional models fails [New test data]}
    \label{probDSA}
\end{figure}

Additionally, model from Ref.\cite{ourYOLOv8}, was mainly trained on single instances for missing hole (mh) and partially closed hole (pch) defects. During inference, model was not able to correctly predict above two defect classes if multiple of them are located near each other, as shown in fig. \ref{probDSA}. The model either miss-classified them as closed patches or could not detect them at all. These SEM images (with 2/3 missing holes/partially closed holes) are part of a new test dataset of HEXCH DSA patterns (models never trained with this dataset and only used during inference).Our goal is to demonstrate if proposed SAHI-based ADCD framework can enable appropriate detection and classification of these missed defect instances without re-training.

\begin{table}[]
\renewcommand{\arraystretch}{1.3}
\caption{\textsc{mAP and mAR metrics on LS (ADI) validation dataset for all considered model variants, using standard inference. Thresholds used are 0.5 for IoU, and 0.25 for confidence score. Best mAP for each YOLO version in \textbf{BOLD}.}}
\label{adiresultss}
\centering
\begin{tabular}{|c||c||c|}
\hline
\textbf{Model} & \textbf{mAP}  &\textbf{mAR} \\
\hline
\textit{YOLOv8n} & 0.827 & 0.869 \\ \hline
\textit{YOLOv8s} & 0.833 & 0.88 \\ \hline
\textit{YOLOv8m} & \textbf{0.856} & 0.904 \\ \hline
\textit{YOLOv8l} & 0.84 & 0.883 \\ \hline
\textit{YOLOv8x} & 0.834 & 0.882 \\ \hline
\textit{YOLOv5n} & 0.85 & 0.901 \\ \hline
\textit{YOLOv5m} & 0.822 & 0.863 \\ \hline
\textit{YOLOv5x} & \textbf{0.856} & 0.893 \\ \hline
\end{tabular}
\end{table}

\begin{table}[]
\renewcommand{\arraystretch}{1.3}
\caption{\textsc{mAP and mAR metrics on HEXCH validation dataset for all considered model variants, using standard inference. Thresholds used are 0.5 for IoU, and 0.25 for confidence score. Best mAP for each YOLO version in \textbf{BOLD}.}}
\label{dsaresultss}
\centering
\begin{tabular}{|c||c||c|}
\hline
\textbf{Model} & \textbf{mAP}  &\textbf{mAR} \\
\hline
\textit{YOLOv8n} & 0.842 & 0.875 \\ \hline
\textit{YOLOv8s} & 0.882 & 0.896 \\ \hline
\textit{YOLOv8m} & 0.865 & 0.87 \\ \hline
\textit{YOLOv8l} & 0.867 & 0.87 \\ \hline
\textit{YOLOv8x} & \textbf{0.884} & 0.891 \\ \hline
\textit{YOLOv5n} & \textbf{0.887} & 0.896 \\ \hline
\textit{YOLOv5m} & 0.866 & 0.891 \\ \hline
\textit{YOLOv5x} & 0.86 & 0.87 \\ \hline
\end{tabular}
\end{table}

\begin{figure}[h]
  \centering
    \begin{subfigure}{0.49\textwidth}
    \includegraphics[width=\linewidth]{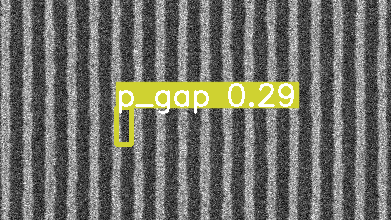}
  \end{subfigure}
  \begin{subfigure}{0.49\textwidth}
    \includegraphics[width=\linewidth]{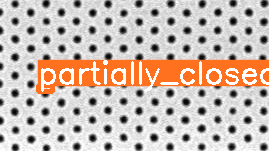}
\end{subfigure}
  \caption{Examples of false-positive predictions caused by predictions at slice edge for pgap (left) and pch (right)}
    \label{ex_fp}
\end{figure}

\subsection{Refining Predictions}\label{refine}
\begin{figure}[b]
    \centering
      \begin{subfigure}{0.49\textwidth}
    \includegraphics[width=\linewidth]{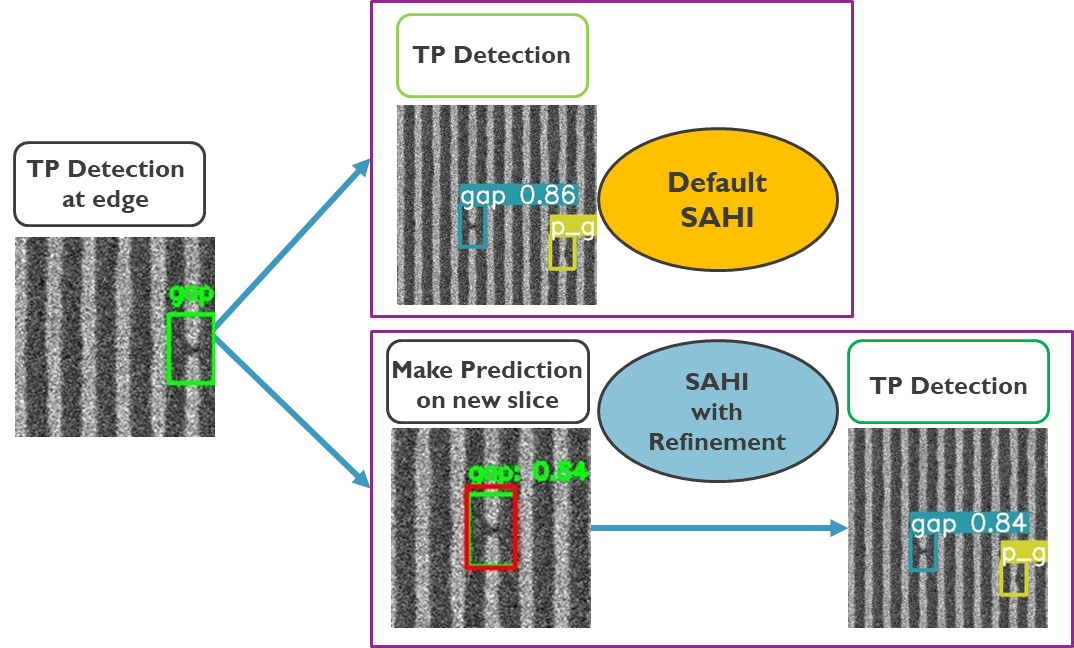}
  \end{subfigure}
       \begin{subfigure}{0.49\textwidth}
    \includegraphics[width=\linewidth]{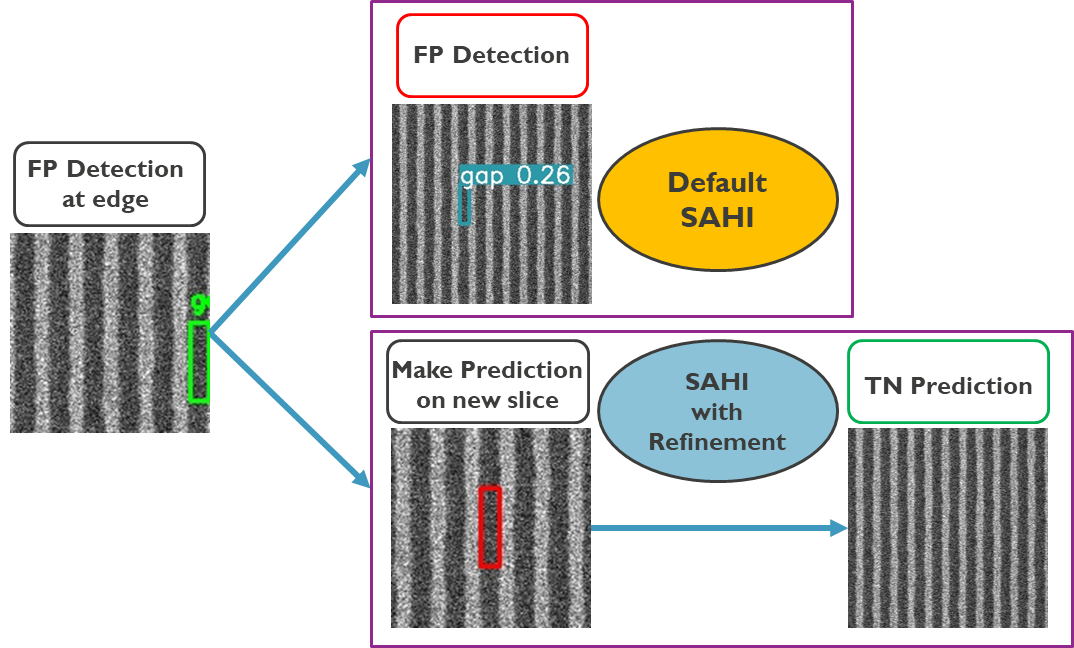}

  \end{subfigure}
      \caption{Our proposed refinement strategy for SAHI framework. Illustrated for case of TP prediction (left) and FP prediction (right) at slice edge. Red bounding box is detection at edge. When set threshold IoU between Red and bounding boxes of predictions on new slice (Green) is not reached, predictions are discarded.}
    \label{refine_strat}
\end{figure}
In initial experimentation, it was found that SAHI caused a new type of false-positive, related to detections made at the edge of a slice, as demonstrated in fig.\ref{ex_fp}. While setting a higher confidence threshold can eliminate these, it also eliminates a lot of true-positive predictions. To maximize the number of true-positive predictions while eliminating false-positives due to edges of the slice, we propose an extension of the SAHI framework. In normal SAHI, the image is sliced, and final predictions result from postprocessing of predictions on each slice. As an extension of this, we propose to keep track of all predictions with bounding boxes ending or starting at an edge of the considered slice. Once all predictions have been made and postprocessed, a new slice is made for each such prediction, with its bounding box at the center of the slice. These slices are now passed again to the model. If a prediction is made that surpasses the predefined IoU threshold with the original bounding box, it is saved as a prediction. Otherwise, it is discarded. Our proposed refinement strategy is illustrated in fig.\ref{refine_strat}. Optionally, multiple models can be used for making predictions on the new slices, and affirmative, consensus, or unanimous voting can be used to confirm if the edge prediction is valid. For validating this strategy, results using SAHI strategy with and without refinement are compared on each dataset for the best performing model at slice size 128.
\section{Results}

\subsection{Line-Space Dataset}\label{adi_res}
\begin{figure}[h]
  \centering
  \begin{subfigure}[t]{0.32\textwidth}
    \includegraphics[width=\linewidth]{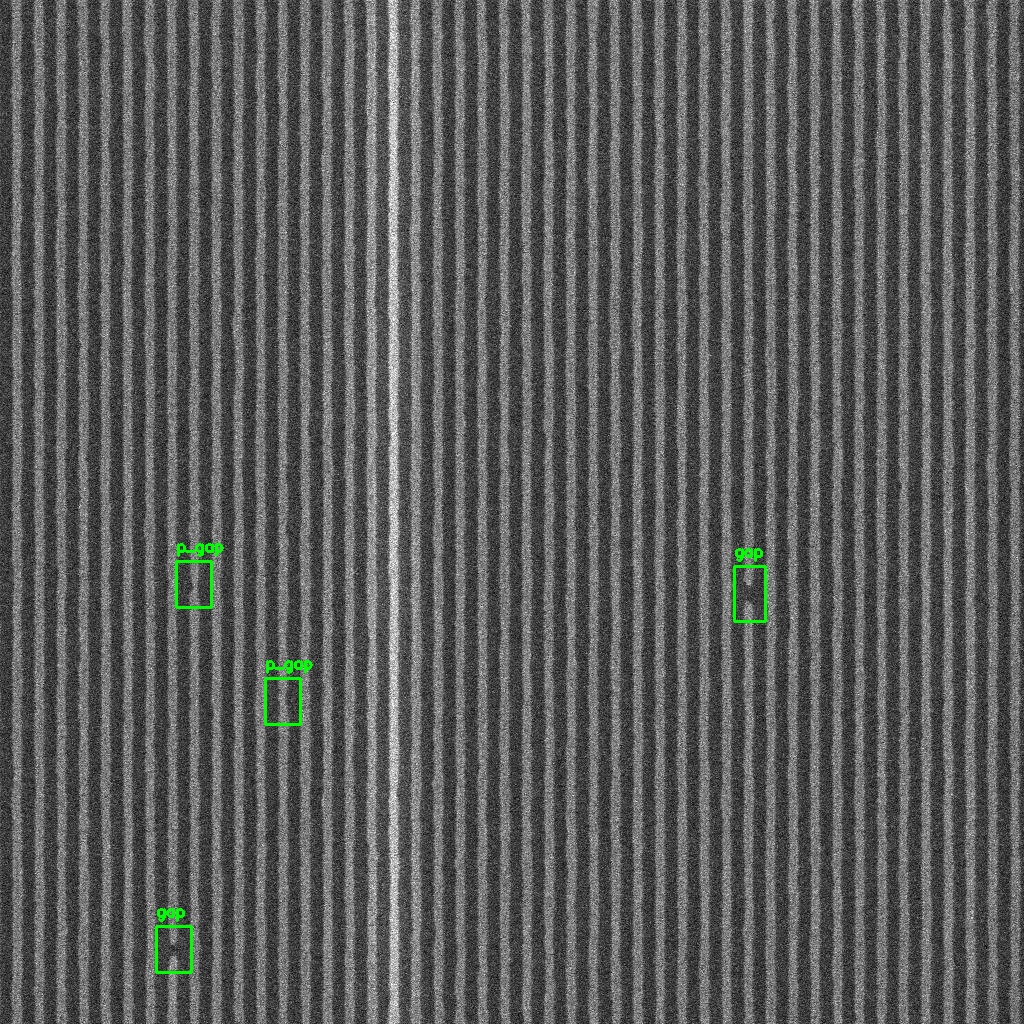}
    \caption{Human Annotation}
  \end{subfigure}
  \begin{subfigure}[t]{0.32\textwidth}
    \includegraphics[width=\linewidth]{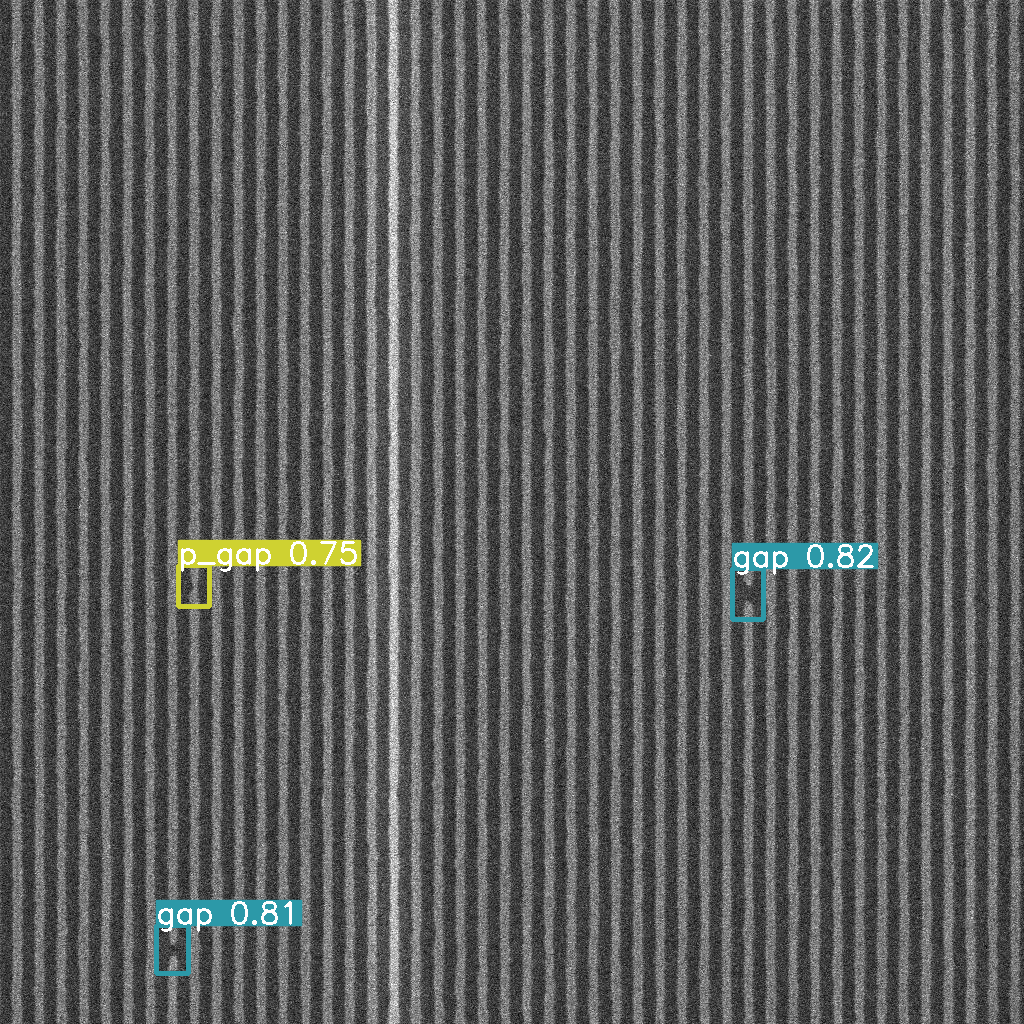}
    \caption{Inference (without SAHI)}
  \end{subfigure}
  \begin{subfigure}[t]{0.32\textwidth}
    \includegraphics[width=\linewidth]{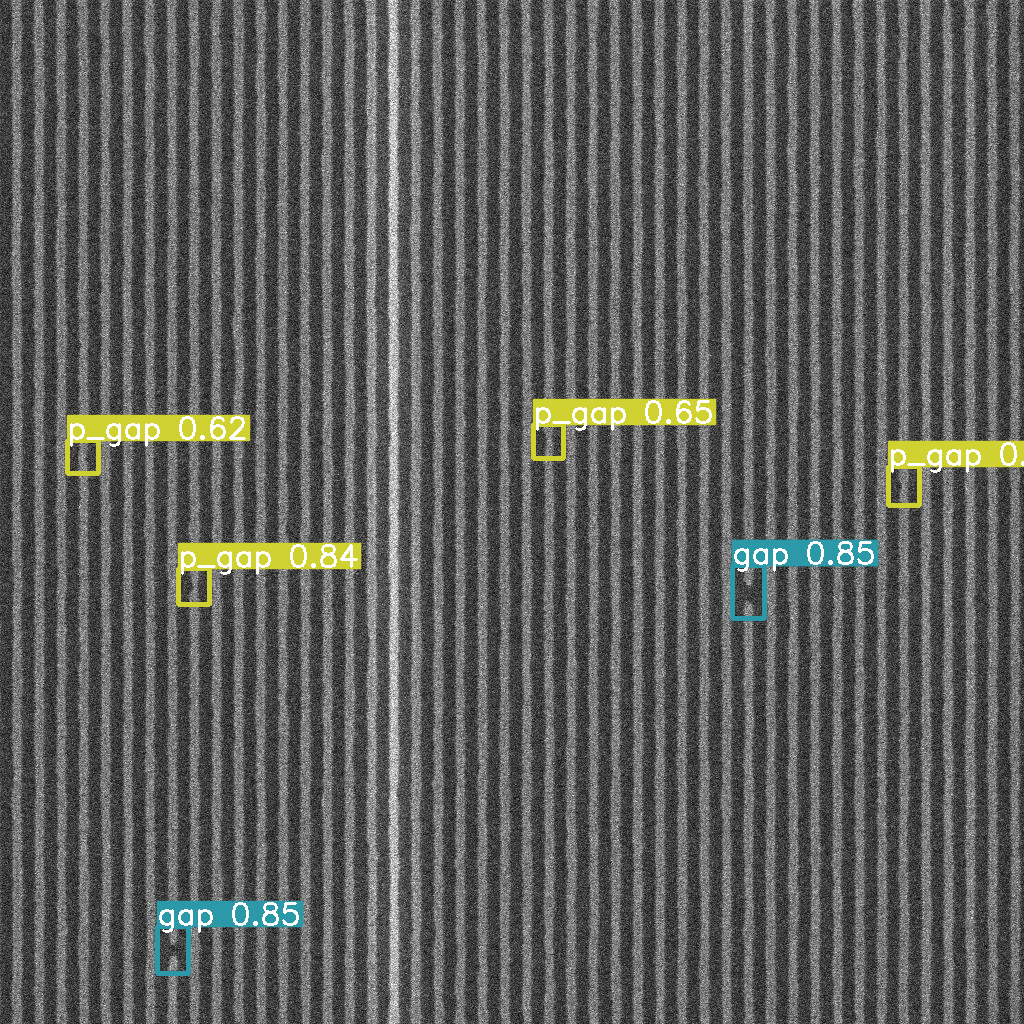}
    \caption{Inference (with SAHI, proposed approach)}
  \end{subfigure}
  \caption{Example defects (a) Annotated by human expert [included missing defects], (b) Predicted by YOLOv8m without SAHI and (c) Predicted by proposed SAHI-enabled ADCD framework with YOLOv8m.}
\end{figure}
\begin{table}[h]
\renewcommand{\arraystretch}{1.3}
\caption{\textsc{Gap and pgap detection results on LS (ADI) dataset for inference without SAHI and with SAHI at various slicing sizes. *Best overall inference results in \textbf{BOLD}.}}
\label{adiresults}
\centering
\begin{tabular}{|c||c||c||c||c||c||c||c|}
\hline
\multirow{2}{*}{\textbf{Model}} & \multicolumn{2}{|c|}{\multirow{2}{*}{\textbf{Inference Strategy}}}  & \multicolumn{2}{|c|}{\textbf{gap}} & \multicolumn{2}{|c|}{\textbf{pgap}} \\
\cline{4-7}
& \multicolumn{2}{|c|}{} &TP & FP & TP & FP \\ 
\hline
\multirow{4}{*}{\textbf{YOLOv5x}}&  \multicolumn{2}{|c|}{Without SAHI} & 173 &1 & 49&0\\
\cline{2-7}
& \multirow{3}{*}{SAHI} &\textbf{128}&\textbf{177} &\textbf{4} &\textbf{173} &\textbf{6}\\
\cline{3-7}
& &256&165 &1 & 69&0\\
\cline{3-7}
& &512&166 & 4&64 &0\\
\hline
\multirow{4}{*}{\textbf{YOLOv8m}}& \multicolumn{2}{|c|}{Without SAHI}& 170 & 0 &62 &0\\
\cline{2-7}
& \multirow{3}{*}{SAHI}&128& 210&9 & 164&10\\
\cline{3-7}
& &256& 163& 3& 67&0\\
\cline{3-7}
& &512& 161& 4&56 &0\\
\hline
\end{tabular}
\end{table}

Table \ref{adiresults} shows the inference results obtained both with and without SAHI using YOLOv8m and YOLOv5x models, which were the best performing variants during standard training and validation. While SAHI-based inference for slicing sizes of 256 or 512 does not offer significant improvements in number of true-positive predictions compared to conventional inference, slicing size of 128 significantly increases the true-positive count of predicted pgap instances by more than \textbf{100}. SAHI-based inference detects additional new defect instances against human annotations, as demonstrated in fig. 7. Interestingly, SAHI-based inference does not cause significant increases in true-positive predictions for gap when using YOLOv5x model, but increases it significantly when YOLOv8m is used. 
However, for all slice sizes, SAHI-based inference increases the number of false-positive predictions. These are caused primarily as an artifact of slicing, as discussed in section \ref{refine}. When the model makes predictions on the slice, it may only have partial information on the lines at the slice edges. Due to this, it sometimes predicts a defect on that line, while if the entire width of the line is analysed, no defect is present. An example of such a false-positive prediction is demonstrated in fig.\ref{ex_fp}.

\subsection{HEXCH DSA Dataset}\label{dsa_res}
\begin{figure}[h]
  \centering
  \begin{subfigure}[t]{0.32\textwidth}
    \includegraphics[width=\linewidth]{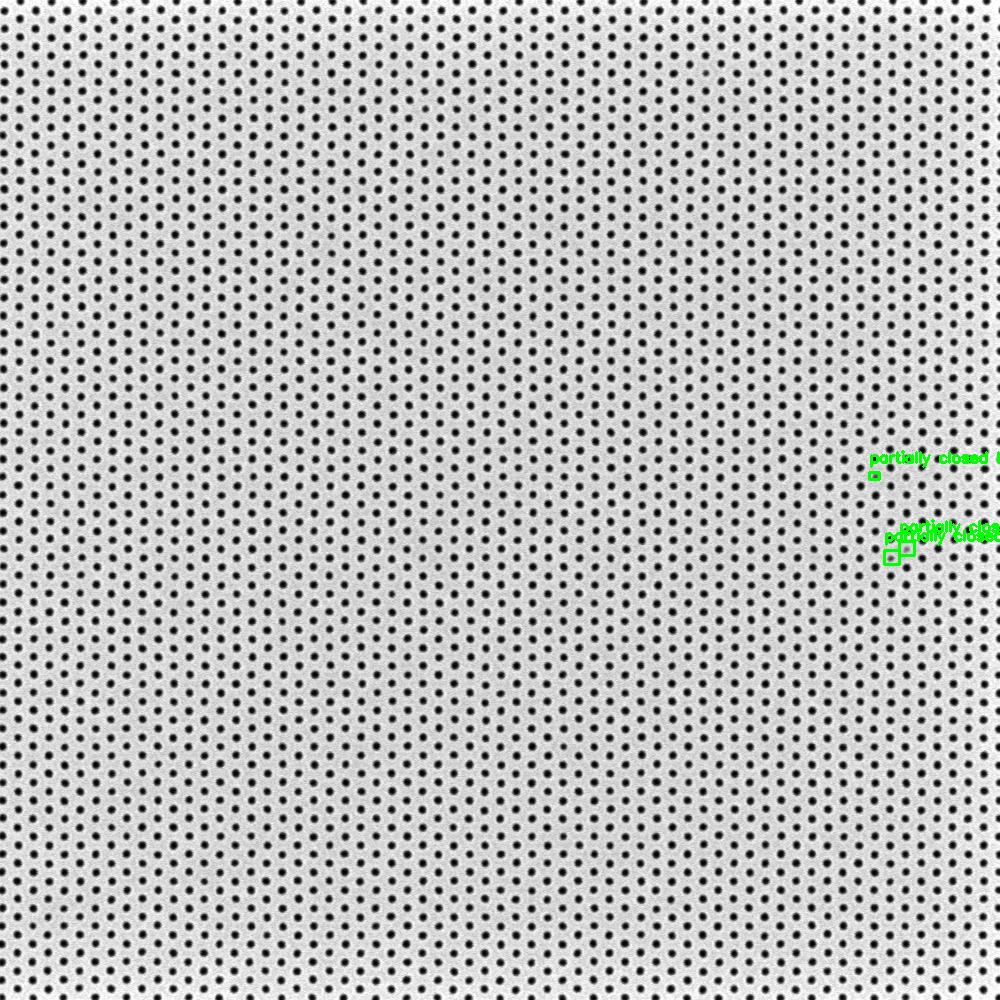}
    \caption{Human Annotation}
  \end{subfigure}
  \begin{subfigure}[t]{0.32\textwidth}
    \includegraphics[width=\linewidth]{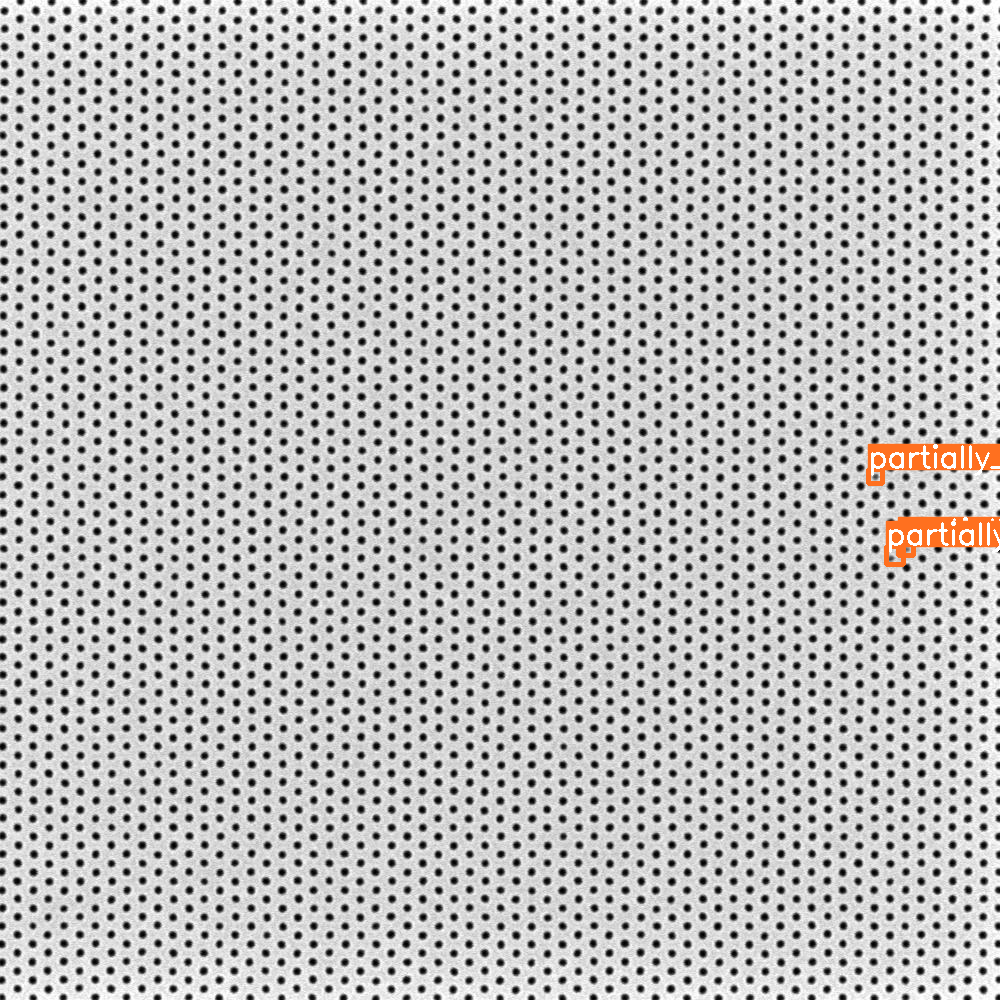}
    \caption{Inference (without SAHI)}
  \end{subfigure}
  \begin{subfigure}[t]{0.32\textwidth}
    \includegraphics[width=\linewidth]{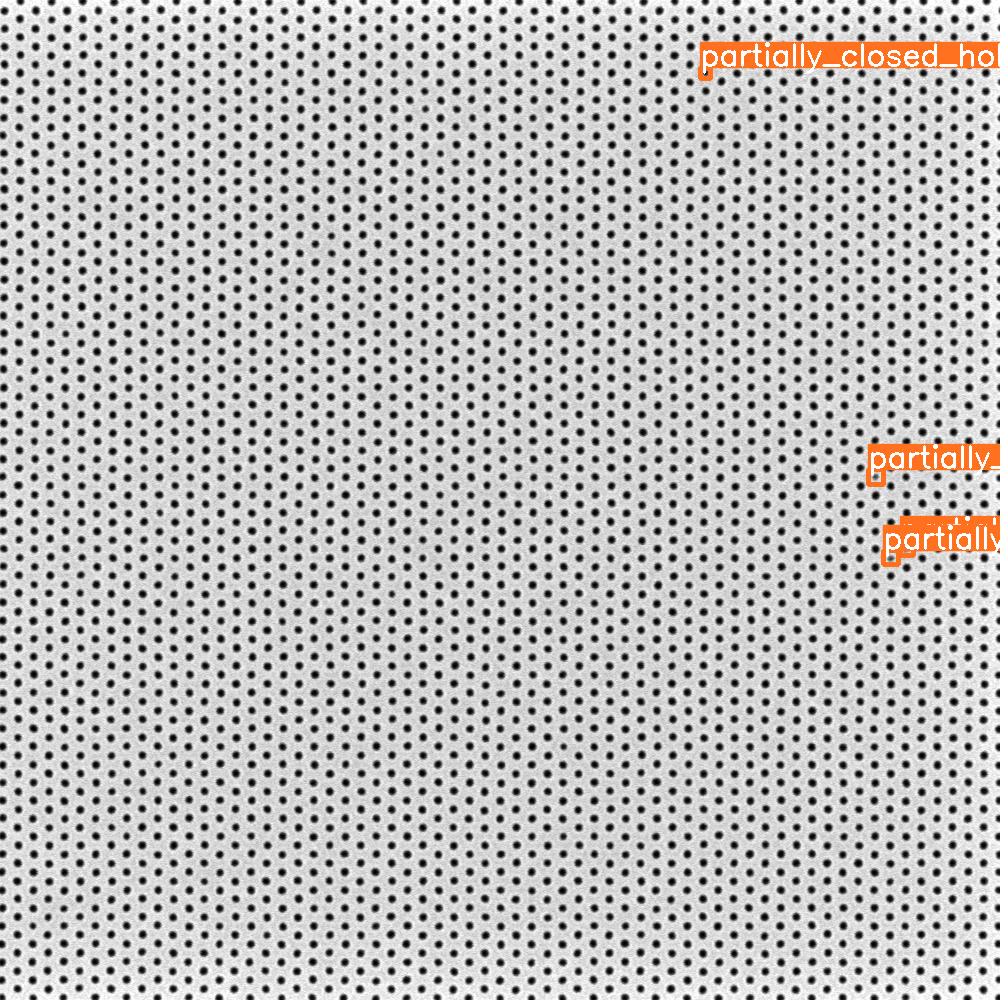}
    \caption{Inference (with SAHI, proposed approach)}
  \end{subfigure}
  \caption{Example defects (a) Annotated by human expert, (b) Predicted by YOLOv5n without SAHI and (c) Predicted by proposed SAHI-enabled ADCD framework with YOLOv5n.}
\end{figure}

\begin{figure}[h]
  \centering
    \begin{subfigure}[t]{0.32\textwidth}
    \includegraphics[width=\linewidth]{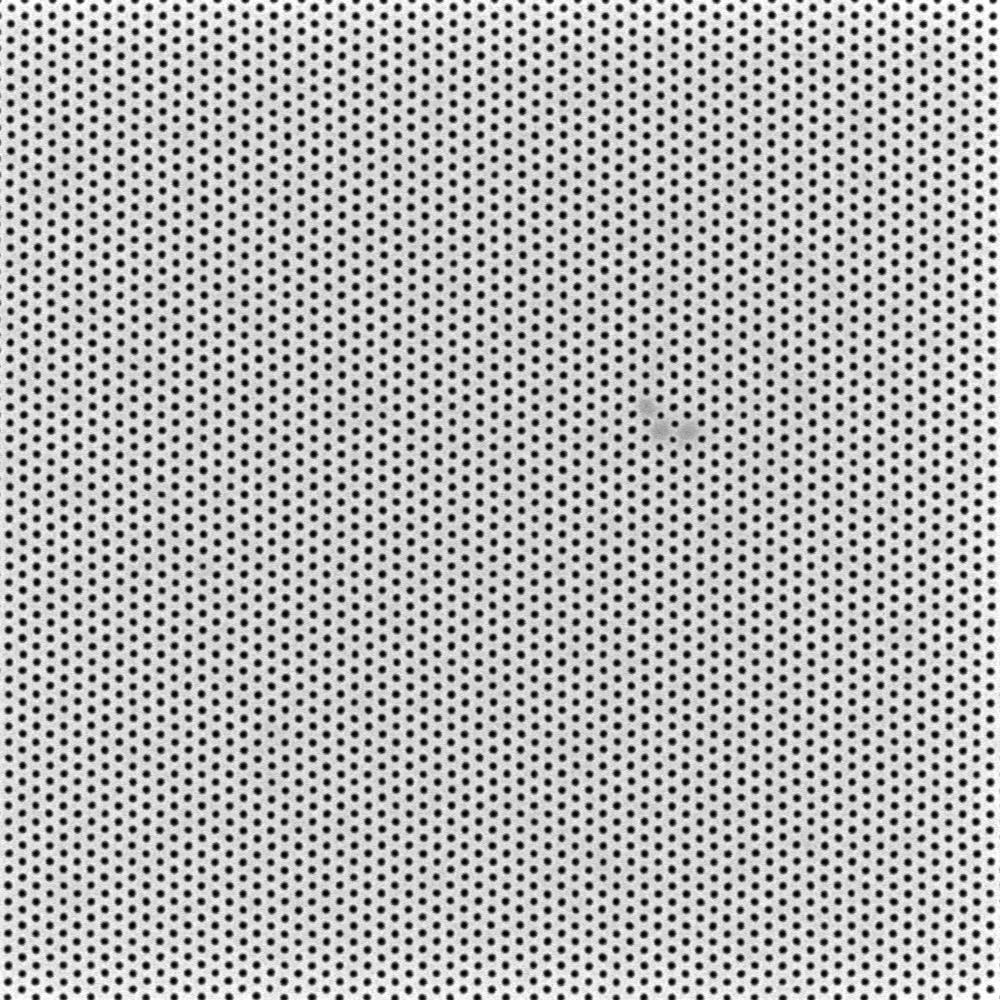}
    \caption{Test image (unannotated)}
  \end{subfigure}
  \begin{subfigure}[t]{0.32\textwidth}
    \includegraphics[width=\linewidth]{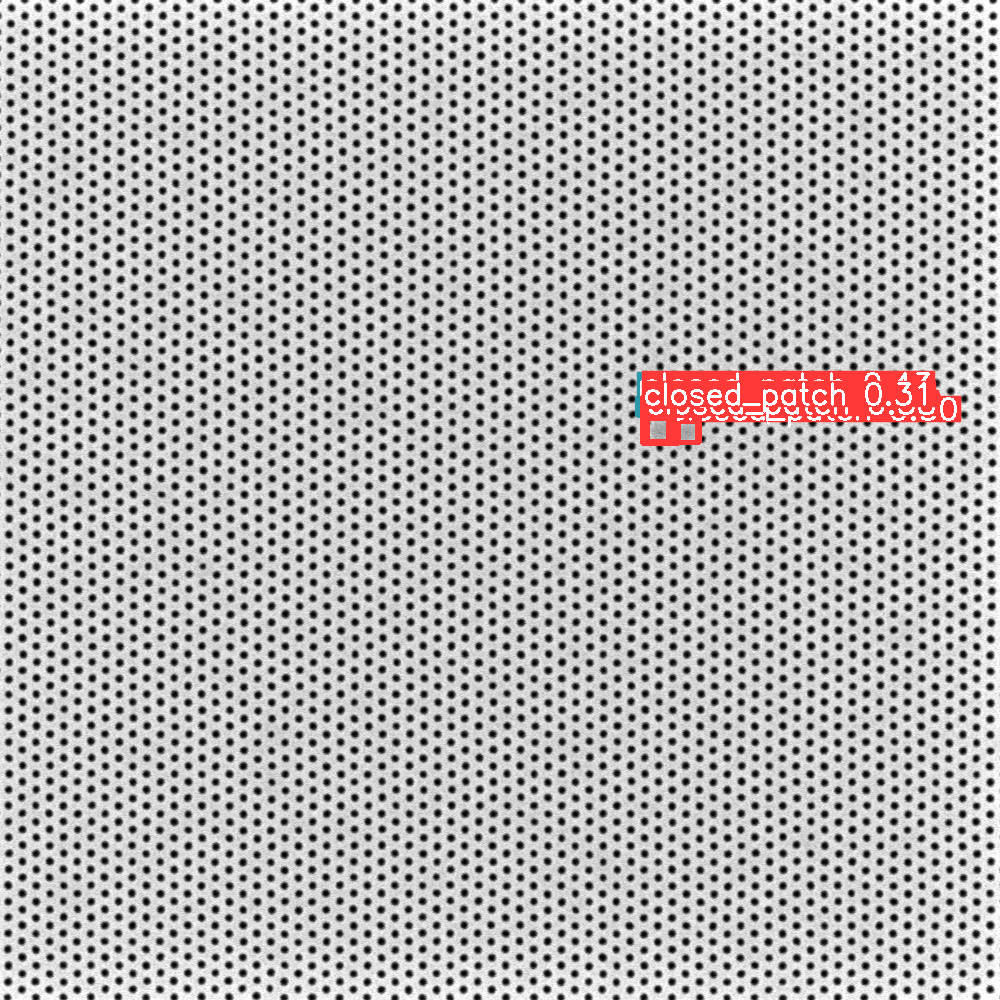}
    \caption{Inference (without SAHI)}
  \end{subfigure}
  \begin{subfigure}[t]{0.32\textwidth}
    \includegraphics[width=\linewidth]{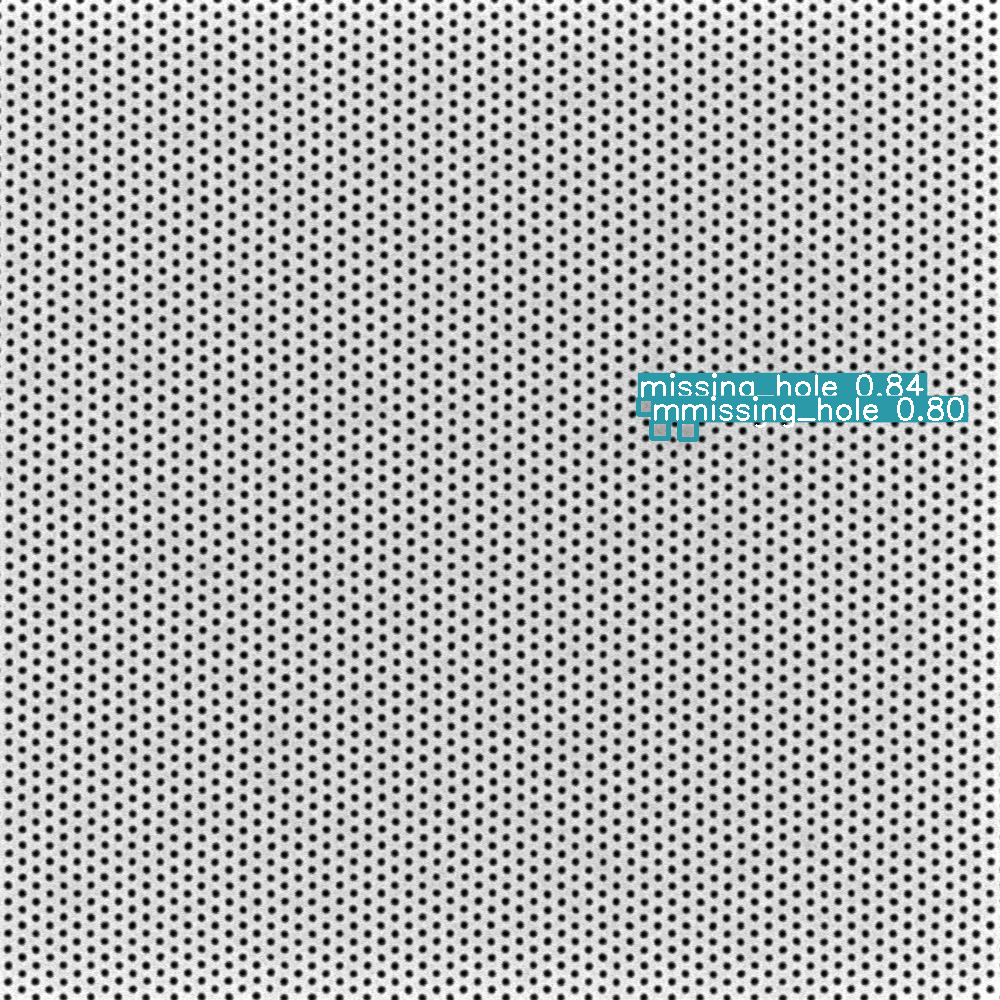}
    \caption{Inference (with SAHI, proposed approach)}
  \end{subfigure}
  \caption{Example defects predicted on HEXCH DSA test data (three missing holes): (a) Test image (unannotated), (b) normal inference, and (c) SAHI-based inference.}
    \label{probdsa_r}
\end{figure}
Table \ref{dsaresults} shows the inference results obtained both with and without SAHI using YOLOv8x and YOLOv5n models, which were the best performing variants during conventional training and inference. In contrast to previously discussed ADI dataset, true-positive predictions are increased significantly not only at 128, but also 256 slicing size. Improvement against both inference without SAHI and human annotation, is shown in fig. 8. However, the number of false-positive predictions for partially closed hole defect type also increases significantly. Again, this may be caused due to impartial information at the slice edges, as shown in fig.\ref{ex_fp}.
Performance comparison on the new test dataset (with 2/3 missing holes/partially closed holes), for both conventional inference method and proposed SAHI-based inference method is demonstrated in fig. \ref{probdsa_r}.  While proposed SAHI-based inference detects and classifies all (previously unseen and untrained) defects correctly, conventional inference method either miss-classifies them as closed patches or background or could not detect them at all.

\begin{table}[]
\renewcommand{\arraystretch}{1.3}
\caption{\textsc{Partially closed and missing hole detection results on HEXCH DSA dataset for inference without SAHI
and with SAHI at various slicing sizes. *Best overall inference results in \textbf{BOLD}.}}
\label{dsaresults}
\centering
\begin{tabular}{|c||c||c||c||c||c||c||c|}
\hline
\multirow{2}{*}{\textbf{Model}} & \multicolumn{2}{|c|}{\multirow{2}{*}{\textbf{Inference Strategy}}}  & \multicolumn{2}{|c|}{\textbf{pch}} & \multicolumn{2}{|c|}{\textbf{mh}} \\
\cline{4-7}
& \multicolumn{2}{|c|}{}& TP & FP & TP & FP \\ 
\hline
\multirow{4}{*}{\textbf{YOLOv5n}}& \multicolumn{2}{|c|}{Without SAHI}& 17 & 0 &7 &0\\
\cline{2-7}
& \multirow{2}{*}{SAHI}&128& 43&122 & 7&1\\
\cline{3-7}
& &256& 30& 0& 9&0\\
\hline
\multirow{4}{*}{\textbf{YOLOv8x}}& \multicolumn{2}{|c|}{Without SAHI} &20 & 0 &7 & 0\\
\cline{2-7}
& \multirow{2}{*}{SAHI}&\textbf{128}&\textbf{47} &\textbf{40} &\textbf{14} &\textbf{0}\\
\cline{3-7}
& &256& 35&11 &8 &0\\
\hline
\end{tabular}
\end{table}

\subsection{SAHI-based ADCD framework with refinement strategy}
Finally, table \ref{sahir_results} demonstrates the results for SAHI-based inference with refinement strategy at slicing size 128 on ADI data with YOLOv5x and on DSA with YOLOv8x. The proposed approach eliminates nearly all false-positive predictions on both datasets, properly validating it. On HEXCH DSA dataset, nearly no significant reduction in true-positive prediction numbers is found, while for LS (ADI) dataset, this number is significantly reduced (while compared against expert human annotation and conventional inference, true-positive predictions for pgap are still improved by $\sim$x2). Further to be investigated, if pgap has more ambiguous learn-able features compared to other defect instances (for HEXCH DSA dataset), and thus model may not always be conscious of the defect on the refined slice, in the same way experts may not always agree on whether it is a pgap.
\begin{table}[h]
\renewcommand{\arraystretch}{1.3}
\caption{\textsc{Improved defect detection results for both datasets with SAHI-based ADCD framework with refinement strategy at slicing size of 128.}}
\label{sahir_results}
\centering
\begin{tabular}{|c||c||c||c||c||c||c|}
\hline
\multirow{1}{*}{\textbf{Model}}   & \textbf{Defect Type}& \textbf{TP} & \textbf{FP} & \textbf{FP Reduction by Refinement}&\textbf{Human Annotation} \\
\hline
\multirow{2}{*}{\textbf{YOLOv5x}}& \textit{gap}&175 & 0&100\%&156\\
\cline{2-6}
&\textit{pgap}&157 &1 &83.3\%& 49 \\
\hline
\multirow{2}{*}{\textbf{YOLOv8x}}&\textit{pch} &46 & 2 & 95\%&16\\
\cline{2-6}
&\textit{mh} &11 & 0 &n/a& 8\\
\hline
\end{tabular}
\end{table}

\subsection{Comparison of metrics using different ground truth annotations}
In the previous sections, it has been established that using SAHI, many defect instances are detected which were not annotated by humans (manual labeling counts demonstrated in tables \ref{adidataset},\ref{tdataset} against TP predictions in table \ref{sahir_results}, specifically for most challenging defect instances to label as pgap, pch and mh, respectively). Hence, when AP is calculated against human annotations as ground truth, a lot of correct predictions (TP's) are being considered as FPs, and negatively influence the AP/mAP scores, considering Eqn. 1 and Eqn. 2 for P (Precision) and R (Recall) as:
\begin{equation}
    P := \frac{Cumulative\: TP}{ Cumulative\:TP + Cumulative\:FP }
\end{equation}, 
\begin{equation}
    R := \frac{Cumulative\: TP} {Total \:Ground \:Truths}
\end{equation}

To study the effect this has on AP (Average Precision) and AR (Average Recall) metrics, we compared predictions (of detections and classifications) of best performing models (YOLOv8m for LS (ADI) dataset, and YOLOv5n for HEXCH dataset) -without and -with SAHI against three ground-truth annotation/labeling strategies, towards calculating these two metrics. First, human annotation is used as ground truth. Second, predictions made by selected YOLO model without SAHI are used as ground truth in calculating AP and AR. Finally, predictions from same YOLO model with SAHI are used as ground truth. YOLOv5x (-without and -with SAHI) model predictions are used as ground-truth for LS (ADI) dataset and YOLOv8x (-without and -with SAHI) model predictions are used as ground-truth for HEXCH dataset. Results are shown in tables \ref{AP_ADI} and \ref{AP_DSA}.
Compared to AP score using human annotations as ground truth, with YOLO (-without or -with SAHI) based predictions as ground-truth, SAHI-based inference methods achieved relatively better AP/AR metrics on probable gap, missing hole and partially closed hole defect instances. This means that a significant number of defect instances have been predicted which were missed in human annotation, but have been predicted by different model variants and inference methods. The major difference between human annotations and average predictions of multiple models (as it cannot be guaranteed as all architecture variants individually will detect all/same instances), and consistency in predictions among various architecture variants and inference methods suggests a high likelihood of the presence of defect instances.
Future work can be extended towards exploring the implementation of ensemble strategies for SAHI-based predictions across different architectures, aiming to minimize or eliminate uncertainties in true positive predictions.
\begin{table}[h]
\renewcommand{\arraystretch}{1.3}
\caption{\textsc{Metrics for LS (ADI) defect types with YOLOv8m [-without and -with SAHI] predictions, calculated and compared against three ground truth annotation/labeling strategies. Y5x stands for YOLOv5x.}}
\label{AP_ADI}
\centering
\begin{tabular}{|c||c||c||c||c||c|}
\hline
\textbf{Detection at Inference} & \textbf{Annotation} & \textbf{gap AP50} & \textbf{gap AR50} & \textbf{pgap AP50} & \textbf{pgap AR50} \\ \hline
\multirow{3}{*}{YOLOv8m without SAHI} & Human & 0.949 & 0.974 & 0.479 & 0.633 \\ \cline{2-6}
 & Y5x without SAHI & 0.949 & 0.957 & 0.806 & 0.849 \\ \cline{2-6}
 & Y5x with SAHI & 0.874 & 0.889 & 0.245 & 0.293 \\ \hline
\multirow{3}{*}{YOLOv8m with SAHI} & Human & 0.95 & 0.974 & 0.346 & 0.714 \\ \cline{2-6}
 & Y5x without SAHI & 0.902 & 0.914 & 0.438 & 0.811 \\ \cline{2-6}
 & Y5x with SAHI & 0.9 & 0.91 & 0.486 & 0.604 \\ \hline
\end{tabular}
\end{table}

\begin{table}[h]
\renewcommand{\arraystretch}{1.3}
\caption{\textsc{Metrics for HEXCH defect types with YOLOv5n [-without and -with SAHI] predictions,  calculated and compared against three ground truth annotation/labeling strategies. Y8x stands for YOLOv8x.}}
\label{AP_DSA}
\centering
\begin{tabular}{|c||c||c||c||c||c|}
\hline
\textbf{Detection at Inference} & \textbf{Annotation} & \textbf{pch AP50} & \textbf{pch AR50} & \textbf{mh AP50} & \textbf{mh AR50} \\ \hline
\multirow{3}{*}{YOLOv5n without SAHI} & Human & 0.808 & 0.812 & 0.871 & 0.875 \\ \cline{2-6}
 & Y8x without SAHI & 0.74 & 0.75 & 1 & 1 \\ \cline{2-6}
 & Y8x with SAHI & 0.312 & 0.312 & 0.634 & 0.636 \\ \hline
\multirow{3}{*}{YOLOv5n with SAHI} & Human & 0.531 & 0.875 & 0.871 & 0.875 \\ \cline{2-6}
 & Y8x without SAHI & 0.783 & 0.9 & 1 & 1 \\ \cline{2-6}
 & Y8x with SAHI & 0.772 & 0.792 & 0.634 & 0.636 \\ \hline
\end{tabular}
\end{table}

\section{Discussion and Future Work}
The results demonstrated in this research work not only show a significant improvement in defect inspection performance using SAHI, but also highlight that manual annotations (by human experts) may also fail to capture all defect instances present in those SEM images. For example, in HEXCH DSA validation dataset, only 16 partially closed holes were annotated. In contrast, proposed SAHI-based inference method detected upwards of 47 (using, illustrating the fact that at least 31 pch's were not annotated. For LS (ADI) dataset, the same pattern can be found for pgap, and in lesser terms for gap.
It is very likely that significant number of missed/unannotated defect instances exist in training data. Thus, future work could be directed towards (re-)training objector detector models with previously annotated dataset as well as SAHI-inference-based improved annotations (mostly with newly detected defect instances probably missed during tedious manual annotation). This can be thought of as an alternative data augmentation strategy (for similar datasets), which may result in better training annotations as well as improved defect detection performance.

Finally, while this research work aims to demonstrate first-time application of SAHI in semiconductor defect inspection, performance may be further improved through use of ensembles of different machine learning models and tuning SAHI parameters such as slice overlap, or area threshold during refinement.
\section{Conclusion}
In this research work, SAHI object detection inference framework has been applied to semiconductor defect inspection task. SAHI was investigated for various configurations and two different YOLO model/architecture variants (YOLOv5 and YOLOv8) on two semiconductor SEM datasets: Line-space and Hexagonal Contact-Hole Arrays. Predictions made by these models were further manually evaluated as either true-positives or false-positives for each model and SAHI configuration. On both datasets, application of SAHI framework caused ~2x increase in true-positive predictions for some of the most challenging nano-scale defect types (probable gap and partially closed hole) against mislabeled/unannotated instances. The SAHI-enabled predictions were not only compared against predictions by same models without SAHI, but also against human expert annotations. This highlighted that models using SAHI framework made many more true-positive predictions (as per extensive manual evaluation) compared to expert human annotation. Thus, human annotations also miss significant numbers of defect instances, which are hence unannotated. Because these unannotated defect instances are also present in training data, we propose as future work using SAHI to improve training labels. Finally, we formulated an extension of the SAHI framework, where a new refinement strategy is added to reduce false-positive predictions at slice edges. Most notably, we demonstrate this proposed refinement strategy reduces false-positive predictions on partially closed hole defect type from 40 to 2, without any significant decreases of true-positive predictions.

\end{document}